\documentclass[lettersize,journal,twoside]{IEEEtran}
\usepackage{amsmath,amsfonts}
\usepackage{algorithmic}
\usepackage{algorithm}
\usepackage{array}
\usepackage{amsmath}
\usepackage{amssymb}
\usepackage{pifont}
\usepackage[caption=false,font=normalsize,labelfont=sf,textfont=sf]{subfig}
\usepackage{textcomp}
\usepackage{stfloats}
\usepackage{url}
\usepackage{verbatim}
\usepackage[dvips]{graphicx}
\usepackage[dvips]{epsfig}
\usepackage{cite}
\usepackage{float}
\usepackage{booktabs}
\usepackage{multirow}
\usepackage{svg}
\usepackage{xspace}
\usepackage{subcaption}
\usepackage{overpic}
\usepackage{xcolor}
\usepackage{colortbl}

\usepackage{bm}
\definecolor{citecolor}{RGB}{66,168,235}
\definecolor{BrickRed}{RGB}{203,65,84}
\definecolor{Thistle}{RGB}{216,191,216}

\usepackage{hyperref}
\hypersetup{colorlinks=true,citecolor=citecolor,linkcolor=BrickRed}

\newcommand{\ie}{{\emph{i.e.}},\xspace}

\usepackage{caption}
\newcolumntype{C}[1]{>{\centering\let\newline\\\arraybackslash\hspace{0pt}}m{#1}}

\usepackage[capitalize]{cleveref}
\crefname{section}{Sec.}{Secs.}
\Crefname{section}{Section}{Sections}
\Crefname{table}{Table}{Tables}
\crefname{table}{Tab.}{Tabs.}

\definecolor{citeblue}{rgb}{0.21,0.49,0.74}
\definecolor{darkpink}{rgb}{0.91, 0.33, 0.5}
\definecolor{mygray}{gray}{.94}
\newcommand{\cmark}{\ding{51}}

\begin{document}

\title{Grounding 3D Scene Affordance From Egocentric Interactions}

    \author{Cuiyu Liu, Wei Zhai, Yuhang Yang, Hongchen Luo, Sen Liang, Yang Cao,~\IEEEmembership{Member,~IEEE}, and Zheng-Jun Zha,~\IEEEmembership{Member,~IEEE}
    
    \thanks{Cuiyu Liu, Wei Zhai, Yuhang Yang, Sen Liang, Yang Cao and Zheng-Jun Zha are at the University of Science and Technology of China, Anhui, China. Hongchen Luo is at the Northeastern University, LiaoNing, China.
    
    (e-mail: \{lcy20010626, yyuhang, liangsen\}@mail.ustc.edu.cn, \{wzhai056, forrest, zhazj\}@ustc.edu.cn, luohongchen@ise.neu.edu.cn)}}



\maketitle

\begin{abstract}
Grounding 3D scene affordance aims to locate interactive regions in 3D environments, which is crucial for embodied agents to interact intelligently with their surroundings. Most existing approaches achieve this by mapping semantics to 3D instances based on static geometric structure and visual appearance. This passive strategy limits the agent’s ability to actively perceive and engage with the environment, making it reliant on predefined semantic instructions. In contrast, humans develop complex interaction skills by observing and imitating how others interact with their surroundings. To empower the model with such abilities, we introduce a novel task: grounding 3D scene affordance from egocentric interactions, where the goal is to identify the corresponding affordance regions in a 3D scene based on an egocentric video of an interaction. This task faces the challenges of spatial complexity and alignment complexity across multiple sources. To address these challenges, we propose the Egocentric Interaction-driven 3D Scene Affordance Grounding (Ego-SAG) framework, which utilizes interaction intent to guide the model in focusing on interaction-relevant sub-regions and aligns affordance features from different sources through a bidirectional query decoder mechanism. Furthermore, we introduce the Egocentric Video-3D Scene Affordance Dataset (VSAD), covering a wide range of common interaction types and diverse 3D environments to support this task. Extensive experiments on VSAD validate both the feasibility of the proposed task and the effectiveness of our approach.
\end{abstract}


\section{Introduction}
\IEEEPARstart{T}{he} concept of ``affordance'' is termed the ``possibilities for interaction'' in an environment by James Gibson \cite{gibson2014ecological}. For embodied agents operating in a 3D space, understanding environmental affordance and identifying interactive regions are fundamental to effective perception, interaction, reasoning, and learning\cite{duan2022survey,wang2024embodiedscan}. Developing this capability is essential for transforming a passive perception system into an intelligent system capable of actively engaging with its environment, which is particularly critical in fields such as embodied intelligence\cite{savva2019habitat}, environmental perception\cite{Xia_2018_CVPR}, action prediction\cite{koppula2013learning,vu2014predicting}, and robot navigation\cite{terblanche2021multimodal,wang2020affordance}. To achieve a higher level of environmental comprehension and interaction, despite reasoning and identifying various types of interactions within a 3D scene, an agent should precisely ground the affordance regions involved in these interactions, enhancing its ability to operate in complex environments.

Recently, increasing attention has been given to understanding affordance in 3D scenes. Among them, one type of approach involves directly mapping semantics to 3D instances based on geometric structure and appearance\cite{he2024segpoint,Delitzas_2024_CVPR,takmaz2023openmask3d}, enabling agents to identify interactive regions within a scene from semantic instructions. While this semantic mapping offers a structured and intuitive way to analyze the environment, it fundamentally depends on passively recognizing and classifying specific objects. Moreover, due to the limitations of predefined semantic concepts, these methods often fail to accurately capture the diverse interaction targets in complex real-world environments, leading to inaccurate affordance perception. Another paradigm leverages reinforcement learning\cite{nagarajan2020learning, kolve2017ai2, 10161431}, where agents interact with simulated 3D environments to learn scene affordance through active exploration. However, this method requires numerous trials to converge, making it inefficient. Additionally, the sim-to-real gap further hinders the effectiveness of these methods.

\begin{figure}[t]
	\centering
	\small
        \begin{overpic}[width=1.\linewidth]{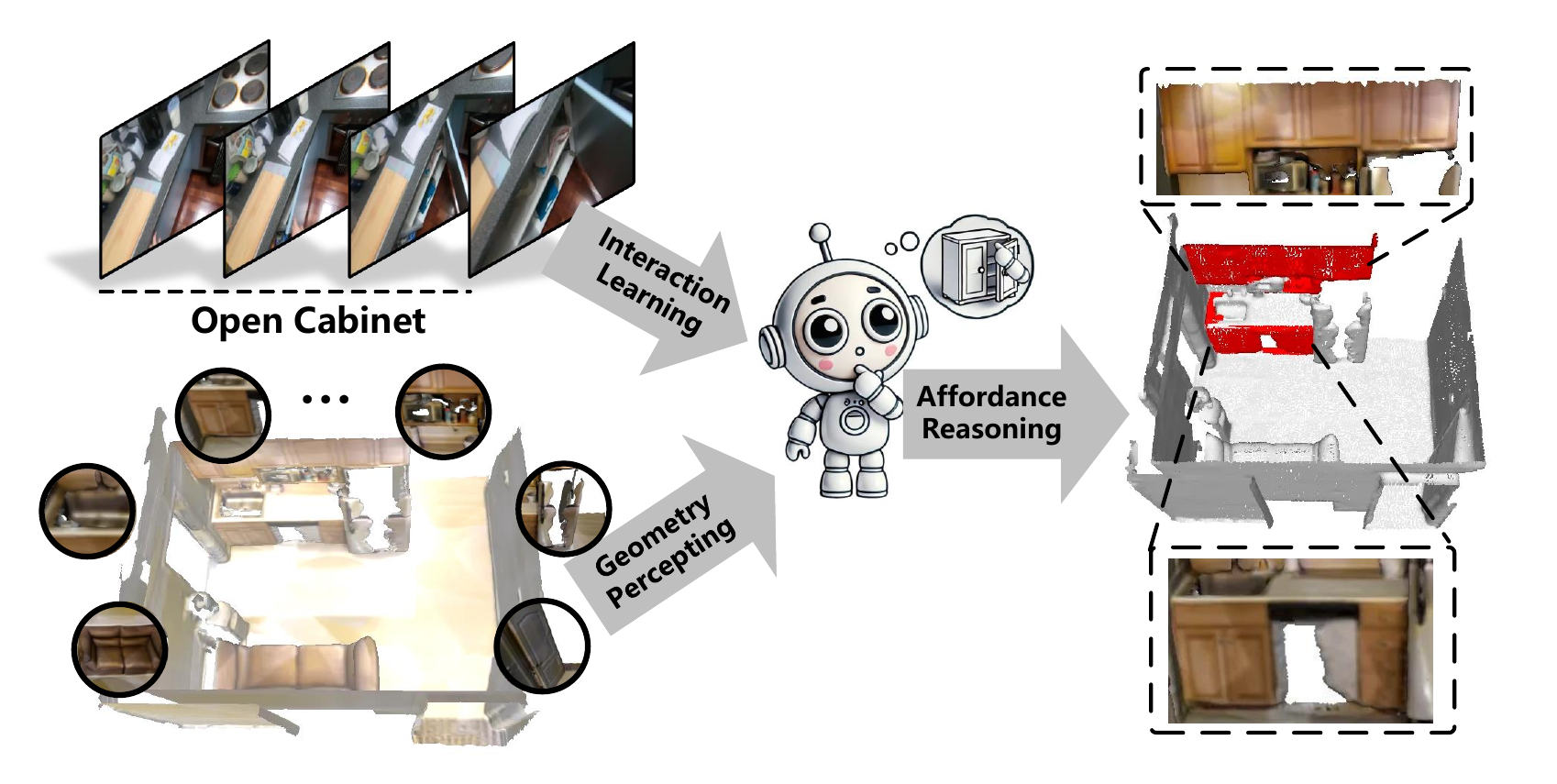}
	\end{overpic}
	\caption{\textbf{Grounding 3D Scene Affordance from Egocentric interactions.} Given an egocentric video describing an interaction and a 3D scene, we propose to ground the corresponding affordance area in the 3D scene.}
 \label{Fig:task}
\end{figure}

These limitations motivate us to explore a new paradigm for understanding 3D scene affordance. In cognitive science, studies show that humans learn complex interaction skills by observing and imitating others' interactions with their surroundings, then applying these skills across different contexts \cite{rizzolatti2004mirror}. The recent availability of affordable head-mounted equipment\cite{apple2023visionpro,gopro,IVUE} and egocentric data offers a more accessible and efficient way to analyze human-environment interactions, making it feasible to construct a model that learns from observations. By focusing on key elements and minimizing redundant visuals, this approach provides richer interaction clues for 3D scene affordance perception. Building on this, we propose grounding 3D scene affordance from egocentric interaction videos, as illustrated in Fig. \ref{Fig:task}, which lies in a more human-like fashion for embodiment simulation, enhancing agents' understanding of 3D spaces by reducing their reliance on reinforcement learning's trial-and-error processes.

To achieve this objective, the primary challenge lies in addressing two key aspects: \textbf{(1) Spatial Complexity:} In 3D environments, the inherent complexity of spatial structures often renders most regions non-essential for interaction, introducing ambiguity in the process of grounding scene affordance. Functional sub-regions within a scene are typically designed based on human interaction patterns and tailored to meet specific functional requirements. For example, kitchen sinks are commonly integrated into countertops to support tasks such as washing. By modeling the relationship between interaction intent observed in videos and the spatial layout of 3D scene sub-regions, it becomes possible to pinpoint the areas most crucial to specific interactions. \textbf{(2) Alignment Complexity:} Variations in user habits, object appearances, and background settings can lead to the same interaction being portrayed differently across videos. At the same time, the corresponding affordance regions within different scenes may also vary significantly in size, position, and structure. These variations create inconsistencies within the feature space, causing complexity in aligning affordance features. However, underlying affordance knowledge shared across these interactions can be harnessed to develop a unified representation that aligns regions with common affordance characteristics.

To tackle these complexities, we propose \textbf{Ego-SAG}, an innovative framework for  \textbf{Ego}centric Interaction-driven 3D \textbf{S}cene \textbf{A}ffordance \textbf{G}rounding, designed to extract and align interaction information from diverse sources to anticipate scene affordance accurately. This framework consists of two integral sub-modules: the \textbf{I}nteraction-Guided \textbf{S}patial \textbf{S}ignificance \textbf{A}llocation Module \textbf{(ISA)} and the \textbf{B}ilateral \textbf{Q}uery \textbf{D}ecoder Module \textbf{(BQD)}. The \textbf{ISA}, designed to handle spatial complexity, is incorporated into each decoding layer of a 3D U-Net. It uses a sampling and grouping strategy to extract features from local sub-regions while modeling the relationship between interaction intent and sub-region layout through the muti-head cross-attention mechanism. This bottom-up approach prioritizes regions most relevant to specific interactions. On the other hand, the \textbf{BQD} constructs a dynamic affordance map across modalities through a bilateral query decoder mechanism, which progressively extracts affordance-related features and optimizes high-dimensional alignment to reveal explicit 3D scene affordance.

Furthermore, existing datasets\cite{dai2017scannet, chang2017matterport3d} fail to fully address the specific requirements of the research task proposed in this paper. To overcome this limitation, we introduce a new dataset named \textbf{V}ideo-3D \textbf{S}cene \textbf{A}ffordance \textbf{D}ataset \textbf{(VSAD)}, which provides an extensive collection of egocentric interaction video-3D scene affordance pairs. \textbf{VSAD} encompasses 17 common affordance categories and 16 different interaction targets, featuring 3,814 egocentric videos and 2,086 3D indoor scenes. These scenes include 7,690 interactive areas that directly correspond to the interaction videos. The dataset covers a wide range of frequently encountered interaction types and various diverse 3D environments, making it highly representative of real-world scenarios. With its comprehensive scope, \textbf{VSAD} serves as a robust and reliable benchmark for both model training and testing.

Our contributions are summarized as follows:
\begin{enumerate}
    \item This paper presents a 3D scene affordance grounding task from the egocentric view and establishes a large-scale \textbf{VSAD} benchmark to facilitate the research for empowering the agent to capture affordance features from egocentric interactions. 
    \item We propose the \textbf{Ego-SAG} framework, which leverages interaction intent to guide the model to focus on interaction-relevant sub-regions within the scene and aligns features of affordance regions between the video and 3D scene through a bidirectional query mechanism, thereby revealing explicit 3D scene affordance.
    \item Experiments conducted on the \textbf{VSAD} dataset demonstrate that \textbf{Ego-SAG} significantly outperforms other representative methods across several related fields and can serve as a strong baseline for future research.
\end{enumerate}
\section{Related Works}
\subsection{Visual Affordance Grounding}
Affordance grounding is crucial for agents to understand their interactions with objects and environments. Various approaches have been explored to achieve visual affordance grounding. Some studies predict object affordance from 2D sources, \ie images and videos \cite{luo2023leverage,zhao2020object,luo2021one,zhai2022one,luo2022learning}, while others utilize natural language to infer the parts of objects where human interaction occurs within 2D data \cite{mi2020intention,lu2022phrase}. However, extending these methods to the 3D physical world remains non-trivial due to the inherent gap between 2D representations and 3D environments. With the introduction of several 3D object datasets, some researchers explore affordance grounding using 3D data. Certain approaches directly map semantic affordance to specific parts of 3D object regions \cite{deng20213d,xu2022partafford,li2024laso}. However, the absence of real-world interaction can hinder the generalization capability, especially when object structures do not correspond neatly to specific affordance. Other methods attempt to infer 3D object affordance and human contact areas through interactions in images\cite{yang2023grounding, yang2024lemon}. However, the limited interaction information in a single image restricts the effectiveness of these methods. While these works mainly focus on single objects, scaling to room-sized environments requires robots to efficiently and effectively explore large-scale 3D spaces for meaningful interactions. Ego-TOPO \cite{nagarajan2020ego} decomposes space into a topological map derived from egocentric activity to obtain environment affordance, but it is constrained to 2D space. Similarly, reinforcement learning approaches, such as \cite{nagarajan2020learning, kolve2017ai2, 10161431}, explore 3D scene affordance, but they are computationally expensive due to the trial-and-error nature of learning. In contrast, unlike prior work, by leveraging the dynamic and contextual interaction information in egocentric videos, our approach enables agents to learn affordance in real-world 3D environments more efficiently, bridging the gap between 2D visual input and 3D scene understanding.
\subsection{Egocentric Video Understanding}
Egocentric vision technology captures human activities from the unique perspective of the camera wearer, providing a distinct action viewpoint \cite{xu2023towards}. With the introduction of several large-scale Egocentric Video datasets in recent years \cite{damen2018scaling,grauman2022ego4d}, this field has rapidly developed, leading to numerous research tasks such as action recognition \cite{zhang2023helping,10184468,8299578}, action prediction \cite{dessalene2021forecasting,zheng2023egocentric}, interactive object prediction \cite{ragusa2023stillfast,liu2020forecasting}, visual question answering \cite{barmann2022did,jia2022egotaskqa}, and interaction intention
anticipation \cite{zhang2024bidirectional,yang2024egochoir,zhang2024pear}. While research on these tasks has significantly advanced machine vision, understanding 2D visual content alone is insufficient for agents operating in the human world to overcome challenges in a 3D environment due to the lack of spatial perception. However, the task proposed in this paper aims to estimate the corresponding fine-grained affordance regions in the 3D environments by identifying and understanding the interactions provided in egocentric videos and the geometric structure in 3D scenes, thereby bridging the gap between 2D and 3D.
\begin{figure*}[t]
	\centering
	\small
        \begin{overpic}[width=1.\linewidth]{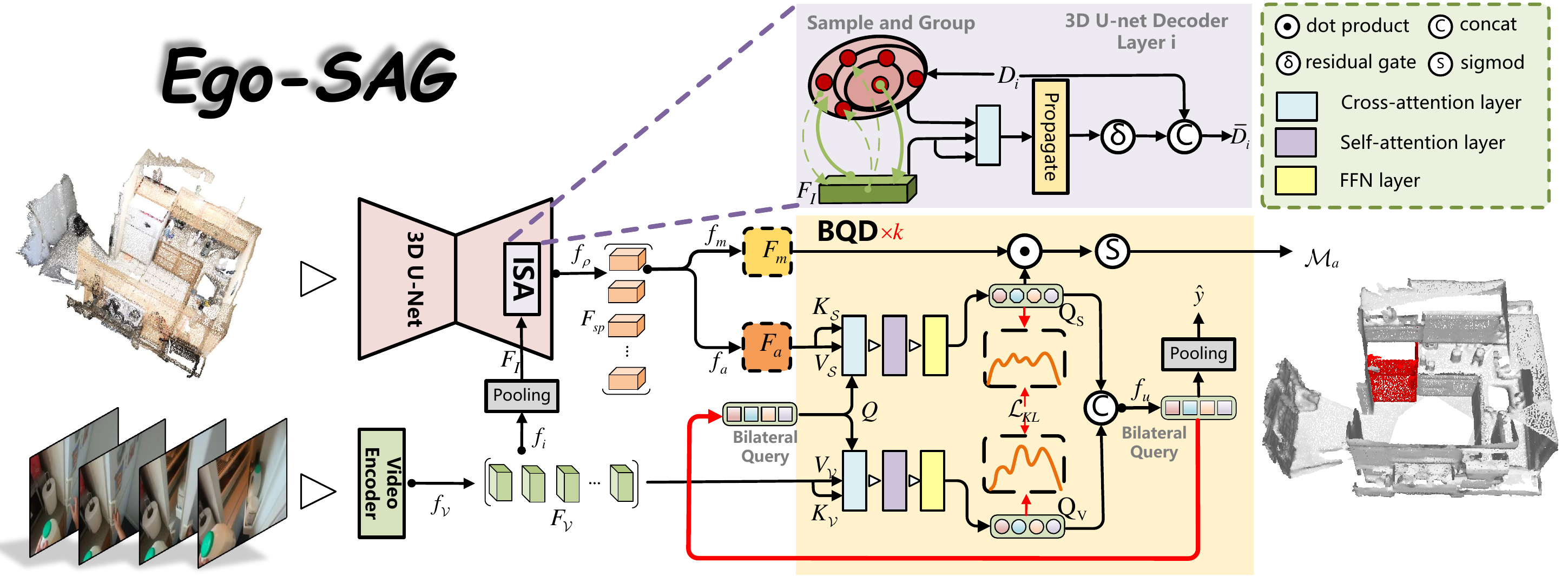}
	\end{overpic}
	\caption{\textbf{Method.} Ego-SAG first uses modality-specific encoders to extract features (Sec.\ref{sec:modalitywise}), modeling the relationship between interaction intent and scene sub-regions during the 3D U-Net decoding stage (Sec.\ref{sec:ISSA}). It then progressively unearths and aligns the corresponding affordance information in the interaction video and the 3D scene (Sec.\ref{BQD}).}
 \label{Fig:method}
\end{figure*}
\subsection{3D Spatial Reasoning}
Recently, research on 3D spatial reasoning has garnered widespread attention, significantly advancing agents' abilities to perceive and comprehend the physical world. Some studies improve agents' understanding of objects within specific semantic categories by segmenting semantics and instances in 3D scenes \cite{lai2023mask,kolodiazhnyi2023oneformer3d,hu2021bidirectional,9913730},  though generalizing to unknown categories remains a challenge. 3D visual grounding tasks \cite{guo2023viewrefer, roh2022languagerefer, chen2023unit3d} further deepen spatial reasoning by using bounding boxes to ground objects in 3D scenes based on text descriptions of spatial relationship or object appearance. However, they lack a focus on interactivity within environments. Others \cite{yang2024llm,ding2024lowis3d,wang2023chat} leverage large language model (LLM) to map rich semantic features to 3D scenes, enabling agents to understand object functions, geometric structures and even answer 3D visual questions. However, these methods primarily focus on static scene perception based on given instructions, overlooking the dynamic interactions that scenes afford. In this paper, we explore a novel approach by grounding affordance in 3D scenes through 2D interactive clues, addressing the gap between static perception and dynamic interaction.
\section{Method}
The pipeline of \textbf{Ego-SAG} is shown in Fig. \ref{Fig:method}, including extracting modality-wise features (Sec.\ref{sec:modalitywise}), modeling the intrinsic relationship between interaction intent and sub-region layout features to highlight interaction-relevant areas (Sec.\ref{sec:ISSA}) and achieving feature alignment between modalities to reveal explicit 3D scene affordance (Sec.\ref{BQD}).
\subsection{Preliminaries}
Given a sample $\left\{ \mathcal{S}, \mathcal{V}, y \right\}$
, where $\mathcal{S} \in \mathbb{R}^{N \times 6}$ is a 3D scene represented by point cloud with three-dimensional coordinates $x, y, z$ and reflection characteristics such as color: $r, g, b$, $\mathcal{V} \in \mathbb{R}^{3 \times T \times H \times W}$ represents a egocentric interaction video clip with $T$ frames of size ${H \times W}$, and $y$ is the video affordance category. Our goal is to learn a model $f_{\theta}$ that anticipates the fine-grained affordance mask $\hat{\mathcal{M}}_a \in \mathbb{R}^{N \times 1}$ on the 3D scene that corresponds to the interaction in egocentric video, along with a video affordance logits $\hat{y}$, expressed as $\hat{\mathcal{M}}_a, \hat{y} = f(\mathcal{S},\mathcal{V};\theta)$, where $\theta$ is the parameters.
\subsection{Modality-wise Feature Extraction}
\label{sec:modalitywise}
\textbf{Video Encoder:} To extract features from the input egocentric video, we utilize the video encoder from Lavila\cite{zhao2023learning}. Specifically, it downsamples the input video clip to 16 frames and then passed through TimeSFormer for feature extraction, which is then projected by a video projector $f_{\mathcal{V}}$ to obtain video features $\mathbf{F}_{\mathcal{V}} \in \mathbb{R}^{C \times TH_{1}W_{1}}$, where $C$ denotes the feature dimension, and $H_1, W_1$ represent the height and width.  $\mathbf{F}_{\mathcal{V}}$ is directly utilized as input for subsequent network modules. \textbf{Scene Encoder:}  We voxelize the point cloud for regular input and utilize a U-Net style backbone that employs submanifold sparse convolution (SSC) and sparse convolution (SC) to extract point-wise scene features $\mathbf{F}_{\mathcal{S}} \in \mathbb{R}^{C \times N}$ in a bottom-up manner. Within this framework, we design an Interaction-guided Spatial Significance Allocation Module, integrated into each decoding layer of the 3D U-Net and introduced in the following subsection. 

\subsection{Interaction-Guided Spatial Significance Allocation Module}
\label{sec:ISSA}
In a 3D scene, extensive non-interactive areas adversely impact the efficiency and accuracy of affordance grounding. To mitigate this, the Interaction-guided Spatial Significance Allocation Module \textbf{(ISA)} firstly captures layout features of these sub-regions through a sample-and-group strategy \cite{qi2017pointnet} at each 3D U-Net decoding layer. It then leverages a muti-head cross-attention mechanism to model the intrinsic relationship between interaction intent and sub-region layout features, thereby highlighting interaction-relevant areas and refining the model's capacity for scene understanding.

Specifically, each layer of the 3D U-Net is indicated as $i \in \left\{1,2,3,4,5\right\}$ in a bottom-up direction. Due to the sparse and irregular distribution of point clouds, extracting geometric and semantic features of scene sub-regions is challenging. In response to this, \textbf{ISA} firstly applies Farthest Point Sampling (FPS) to downsample the $i$-th decoding layer feature $\mathbf{D}{i} \in \mathbb{R}^{C_{i} \times  N{i}}$ from $N_{i}$ points to $N_c$, with these $N_c$ points serving as local centroids. Next, k-nearest neighbor (k-NN) is used to select $k$ nearest neighbors within a radius $r$ around each centroid, aggregating across the $k$ dimension to capture the geometric structure features of uniformly distributed sub-regions, denoted as $\mathbf{G}{i}\in \mathbb{R}^{C{i} \times  N_c}$. These features are further refined through an MLP layer. The process is expressed as:
\begin{equation}
\mathbf{G}_{i} = \text{MLP}(\text{k-NN}(\text{FPS}(\mathbf{D}_{i}, N_c), k, \mathbf{D}_{i})).
\end{equation}
We derive the interaction intention in egocentric videos by applying a convolutional projector $f_i$ to $\mathbf{F}_{\mathcal{V}}$, followed by average pooling over $TH_{1}W_{1}$, yielding a unified video-level intention representation $\mathbf{F}_\mathbf{I} \in \mathbb{R}^{C \times 1}$. Subsequently, a muti-head cross-attention layer $f_\eta$ is employed to model the intrinsic correlation between $\mathbf{G}{i}$ and $\mathbf{F}_\mathbf{I}$, formulated as: 
\begin{equation}
\mathbf{F}_{j} = f_\eta(\mathbf{G}_{i}, \mathbf{F}_\mathbf{I}),
\end{equation}
where $\mathbf{F}_{j} \in \mathbb{R}^{C_{i} \times N_c}$ is the joint feature that emphasizes the interaction-significant sub-regions features. In order to restore the resolution of $\mathbf{F}{j}$ to match the $\mathbf{D}{i}$, the features are propagated \cite{qi2017pointnet++} from each sub-region's centroid back to its corresponding points through weighted interpolation, resulting in $\bar{\mathbf{F}_{j}}$. Finally, a residual gate $\delta$ is introduced to learn a weight matrix, which further filters out areas irrelevant to the interaction intent, and then combines these features with the original point cloud features, formulated as $\bar{\mathbf{D}_{i}} = \mathbf{D}_{i} + \delta(\bar{\mathbf{F}_{j}}) \in \mathbb{R}^{C_i \times N_i}$, which is used to as input of next decoding layer.

Moreover, to reduce the computational burden of subsequent operations and enhance the entire network's representational capability, we pre-sample the 3D scene into $M$ superpoints ($SP$) using the method described in \cite{landrieu2018large}. After obtaining the full-resolution feature map $\mathbf{F}_{\mathcal{S}}$ of the scene, point-level features within these superpoints are aggregated into superpoint-level features $\mathbf{F}_{sp} = f_{\rho}(\mathbf{F}_{\mathcal{S}},SP)\in \mathbb{R}^{C \times M}$ through a superpoint pooling layer $f_{\rho}$. The aggregated $\mathbf{F}_{sp}$ is then fed into the Bilateral Query Decoder module to further explore and align affordance-related contexts between $\mathbf{F}_{sp}$ and $\mathbf{F}_{\mathcal{V}}$.

\subsection{Bilateral Query Decoder Module}
\label{BQD}
The vast differences between the visual appearance of objects in videos and their geometry counterparts in 3D scenes lead to substantial divergence in the distribution of video and scene features within feature space. To generate accurate and consistent representations of the 3D scene affordance regions alongside the corresponding interaction in the egocentric video, Bilateral Query Decoder Module $\textbf{(BQD)}$ leverages a bilateral query mechanism that facilitates interaction between the two modalities, progressively unearthing their consistency, thereby achieving effective feature alignment.

Specifically, $\textbf{BQD}$ consists of $L$ basic layers. Before the first layer, $\mathbf{F}_{sp}$ is projected into affordance features $\mathbf{F}_a$ and mask features $\mathbf{F}_m$ using the MLP-based projectors $f_a$ and $f_m$ respectively, enabling separate processing of affordance and mask information in subsequent layers. Meanwhile, $\textbf{BQD}$ randomly initialize a set of learnable bilateral query vectors $\mathbf{\mathcal{Q}}_0 \in \mathbb{R}^{C \times N}$. At the $\ell$-th layer, $\mathbf{\mathcal{Q}}_{\ell}$ captures affordance information from $\mathbf{F}_a$ and $\mathbf{F}_{\mathcal{V}}$ through transformer-based Geometry Cross-Attention block and Interaction Cross-Attention block, respectively. In these blocks, the input $\mathbf{\mathcal{Q}}_{\ell}$ is initially projected to generate the shared query $\mathbf{Q} = \mathbf{\mathcal{Q}}_{\ell}\mathbf{W}_1$. Simultaneously, the features $\mathbf{F}_a$ and $\mathbf{F}_{\mathcal{V}}$ are projected into distinct sets of keys and values, defined as $\mathbf{K/V}_{s} = \mathbf{F}_a\mathbf{W}_{2/3}$ and $\mathbf{K/V}_{v} = \mathbf{F}_{\mathcal{V}}\mathbf{W}_{4/5}$, where $\mathbf{W}_{1 \sim 5}$ denote the projection matrices. Subsequently, cross-attention is applied to aggregate these features, thereby extracting affordance-related contexts, formulated as:
\begin{equation}
    \mathbf{Q}_{s/v} = \sigma(\mathbf{Q}\cdot\mathbf{K}_{s/v}^{T}/\sqrt{d})\cdot\mathbf{V}^{T}_{s/v},
\end{equation}
where $\mathbf{Q} \in \mathbb{R}^{d \times K}, \mathbf{K/V}_s \in \mathbb{R}^{d \times M}, \mathbf{K/V}_v \in \mathbb{R}^{d  \times TH_1W_1}$, $d$ is the dimension of projection. Following is a self-attention layer $\Gamma$ and a feed-forward network to model the intra-connections between affordance clues extracted by ${\mathbf{Q}}_{s/v}$ in respective modalities, the process is expressed as: $\mathbf{Q}_{s/v} = FFN(\Gamma({\mathbf{Q}}_{s/v}))$. Subsequently, $\mathbf{Q}_s$ and $\mathbf{Q}_v$ are merged to update the bilateral query vectors $\mathbf{\mathcal{Q}}_{\ell+1}$ for the next layer, as described by: $\mathbf{\mathcal{Q}}_{\ell+1} = f_m\left( [\mathbf{Q}_s ; \mathbf{Q}_v] \right)$, where $f_m$ is an MLP and $[\cdot; \cdot]$ denotes concatenation. 

To speed up convergence, we incorporate a prediction head at each layer of \textbf{BQD}. In $\ell$th layer, $\mathbf{\mathcal{Q}}_{\ell+1}$ generated for next layer is pooled to computer $\hat{y}$. Leveraging the mask-aware feature $\mathbf{F}_m$, we directly multiply it by $\mathbf{Q}_s$, followed a sigmoid function to generate superpoint mask $\hat{\mathcal{M}}_{sp} \in \mathbb{R}^{M \times 1} \in \left [0,1 \right] $. Furthermore, as accurate affordance predictions typically occupy only a small fraction of the overall 3D scene, with the majority constituting background, it is necessary to rank and filter the prediction results. To this end, we calculate the dense cosine similarity between each refined scene query vector $\mathbf{Q}_s$ and the pooled video feature $\mathbf{F}_{\mathcal{V}}'$,  employing this similarity as a quality score to assess and refine the predicted masks, formulated as:
\begin{equation}
\omega_i = \frac{\mathbf{Q}_{s,i} \cdot \mathbf{\mathbf{F}_{\mathcal{V}}'}}{\|\mathbf{Q}_{s,i}\| \|\mathbf{\mathbf{F}_{\mathcal{V}}'}\|} \in \mathbb{R}^{N \times 1},
\end{equation}
where $\omega_i$ is the quality score of corresponding $i$th superpoint-wise affordance mask $\hat{\mathcal{M}}_{sp,i}$. And point-wise affordance mask $\hat{\mathcal{M}}_{a}$ can be obtained by: $\hat{\mathcal{M}}_{a}$ = $\hat{\mathcal{M}}_{sp}\left[ SP \right]$.

\begin{figure*}[t]
	\centering
	\small
        \begin{overpic}[width=1.\linewidth]{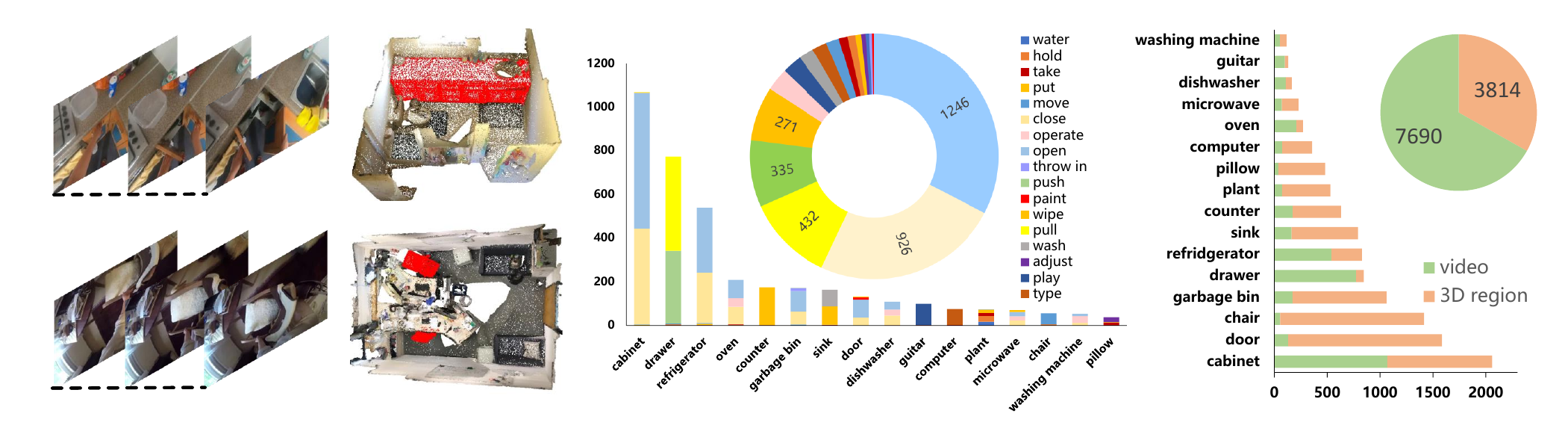}

        \put(20,0){{\textbf{(a)}}}
        \put(52,0){{\textbf{(b)}}}
        \put(88,0){{\textbf{(c)}}}
	\end{overpic}
	\caption{\textbf{Properties of the VSAD Dataset.} \textbf{(a)} Example data pairs from the VSAD dataset, with the video displayed on the left and the corresponding 3D scene visualization on the right. The red regions in the point cloud represent affordance annotations. \textbf{(b)} Distribution of video data: the horizontal axis represents the categories of interactive objects, the vertical axis shows the quantity of data, and different colors indicate various affordance. \textbf{(c)} The ratio of video data to 3D scene affordance regions for each object class. It demonstrates that videos and 3D scenes are not confined to one-to-one pairings, allowing for multiple associations between them.}
 \label{Fig:Dataset}
\end{figure*}

\subsection{Training Objectives}
During training, we formulate ground truth label assignment as an optimal assignment problem. Hungarian matching \cite{kuhn1955hungarian} is employed to pair predictions $\hat{\mathcal{M}}_a$ with GT affordance masks $\mathcal{M}_a$ by minimizing a cost, which consists of a binary cross-entropy (BCE) loss and a dice loss \cite{milletari2016v}, expressed as:
\begin{equation}
\mathcal{C}_{i,j} = \zeta_1 BCE(\hat{\mathcal{M}}_{a,i}, \mathcal{M}_{a,j}) + \zeta_2 Dice(\hat{\mathcal{M}}_{a,i}, \mathcal{M}_{a,j}),
\end{equation}
\begin{equation}
Dice(\hat{\mathcal{M}}_a, \mathcal{M}_a) = 1-2\frac{\hat{\mathcal{M}}_a \cdot \mathcal{M}_a + 1}{\|\ \hat{\mathcal{M}}_a \|\ + \|\ \mathcal{M}_a \|\ + 1},
\end{equation}
where $\mathcal{C}_{i,j}$ represents the cost of matching the $i$th predicted affordance mask with the $j$th GT mask, $\zeta_1$, $\zeta_2$ are hyper-parameters, set to 2 and 5, respectively. 

After the assignment, we compute the final loss between the matched pairs, which comprises four parts: $\mathcal{L}_{CE}$, $\mathcal{L}_{mask}$, $\mathcal{L}_{KL}$, $\mathcal{L}_{con}$. $\mathcal{L}_{CE}$ computes the cross-entropy loss between $y$ and $\hat{y}$, implicitly optimizing the alignment process. To improve the quality of $\hat{\mathcal{M}}_a$, we introduce a $\mathcal{L}_{mask}$, expressed as: $\mathcal{L}_{mask}$ = $l_{bce}$ + $l_{dice}$ + $l_{score}$, where $l_{score} = MSE(\omega, \omega_{gt})$ and the score $\omega_{gt}$ is the IoU between $\hat{\mathcal{M}}_a$ and $\mathcal{M}_a$, used to supervise the quality score of $\hat{\mathcal{M}}_a$. Intuitively, a higher IoU indicates better mask quality. To enable the network focus on the alignment of affordance regions, we apply  KL Divergence (KLD) \cite{bylinskii2018different} to constrain the distribution between $\mathbf{Q}_v$ and $\mathbf{Q}_s$, formulated as: $\mathcal{L}_{KL} = \left( \mathbf{Q}_v \parallel \mathbf{Q}_s \right)$. Since $\mathbf{Q}_v$ reflects the affordance distribution related to video interactions, constraining the distribution of $\mathbf{Q}_s$ with $\mathcal{L}_{KL}$ guides it to focus on the corresponding affordance features in the scene, allowing the alignment and extraction of affordance to optimize mutually. $\mathcal{L}_{con}$ is a contrastive loss that guides the model to learn more discriminative scene affordance feature representations. $\mathbf{Q}_s$ with IoU greater than 0.5 are treated as positive examples, while the rest are considered negative. Given that a scene may contain multiple affordance regions described by interaction video, resulting in multiple positive examples, we adopt a multi-positive example contrastive loss referred to \cite{yuan2021instancerefer}, defined as: 
\begin{equation}
\mathcal{L}_{con} = -\log \frac{\frac{1}{N^+}\sum_{i=1}^{N^+}\exp{({\omega_i}^+})}{\frac{1}{N^+}\sum_{i=1}^{N^+}\exp{({\omega_i}^+)} + \frac{1}{N^-}\sum_{i=1}^{N^-}\exp{({\omega_i}^-})},
\end{equation}
where ${\omega_i}^+$ and ${\omega_i}^-$ represent the scores of positive and negative pairs while $N^+$ and $N^-$ denote the number of positive and negative affordance predictions within a scene. Ultimately, the total loss is formulated as:
\begin{equation}
\mathcal{L}_{total} = \lambda_1\mathcal{L}_{CE} + \lambda_2\mathcal{L}_{mask} + \lambda_3\mathcal{L}_{KL} + \lambda_4\mathcal{L}_{con},
\end{equation}
where $\lambda_1$, $\lambda_2$, $\lambda_3$ and $\lambda_4$ are hyper-parameters to balance the total loss.

During inference, \textbf{Ego-SAG} directly predicts K affordance masks with the highest scores and filters out those with scores less than the threshold $\tau$, which we set as 0.5.
\section{Dataset}

\begin{figure*}[t]
	\centering
	\small
        \begin{overpic}[width=1.\linewidth]{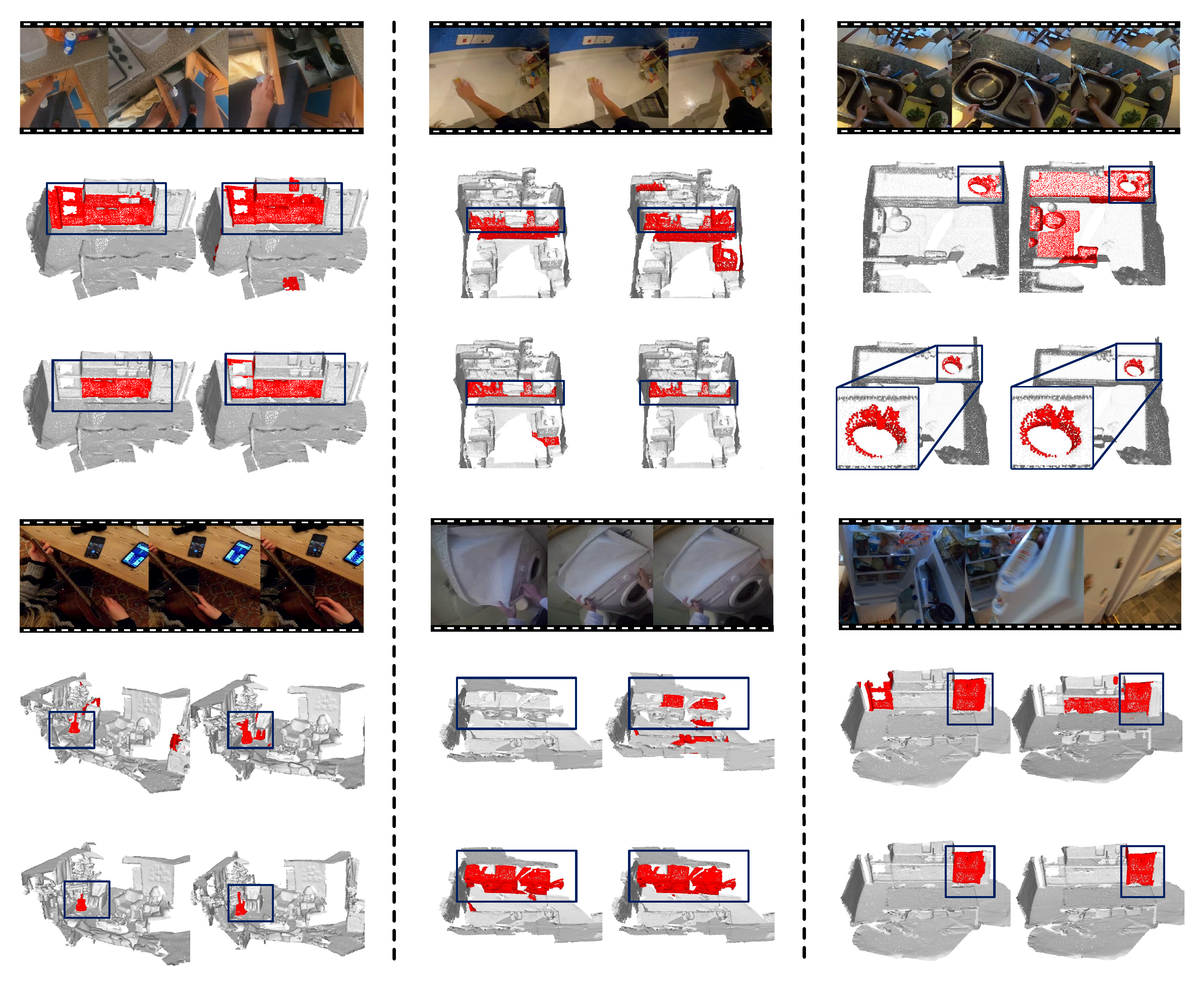}
        \put(9.5,69.5){{\textbf{Cabinet }}}
        \put(17.5,69.5){{\textbf{Open }}}
        \put(5,56){{\textbf{Openins3d}}}
        \put(19,56){{\textbf{Openmask3d}}}
        \put(8,42){{\textbf{Ours}}}
        \put(23,42){{\textbf{GT}}}

        \put(43.5,69.5){{\textbf{Counter}}}
        \put(51.5,69.5){{\textbf{Wipe}}}
        \put(38,56){{\textbf{Openins3d}}}
        \put(52,56){{\textbf{Openmask3d}}}
        \put(41,42){{\textbf{Ours}}}
        \put(56,42){{\textbf{GT}}}

        \put(79.5,69.5){{\textbf{Sink}}}
        \put(84.5,69.5){{\textbf{Wash}}}

        \put(73,56){{\textbf{Openins3d}}}
        \put(86,56){{\textbf{Openmask3d}}}
        \put(75,42){{\textbf{Ours}}}
        \put(89,42){{\textbf{GT}}}

        \put(10,28){{\textbf{Guitar }}}
        \put(17,28){{\textbf{Play }}}

        \put(5,15){{\textbf{Openins3d}}}
        \put(19,15){{\textbf{Openmask3d}}}
        \put(8,1){{\textbf{Ours}}}
        \put(23,1){{\textbf{GT}}}

        \put(39,28){{\textbf{Washing-machine}}}
        \put(54,28){{\textbf{Operate}}}

        \put(38,15){{\textbf{Openins3d}}}
        \put(52,15){{\textbf{Openmask3d}}}
        \put(41,1){{\textbf{Ours}}}
        \put(56,1){{\textbf{GT}}}

        \put(76,28){{\textbf{Refrigerator}}}
        \put(87,28){{\textbf{Close}}}
        \put(73,15){{\textbf{Openins3d}}}
        \put(86,15){{\textbf{Openmask3d}}}
        \put(75,1){{\textbf{Ours}}}
        \put(90,1){{\textbf{GT}}}

	\end{overpic}
	\caption{\textbf{Visualization Results.} Each sample consists of three rows: the first row is the egocentric video demonstrating the interaction that the scene can afford, the second row shows the results of the comparison methods, and the third row shows the result of our method and GT. Scene affordance masks are colored red. Please zoom in for a better visualization.}
 \label{Fig:result}
\end{figure*}
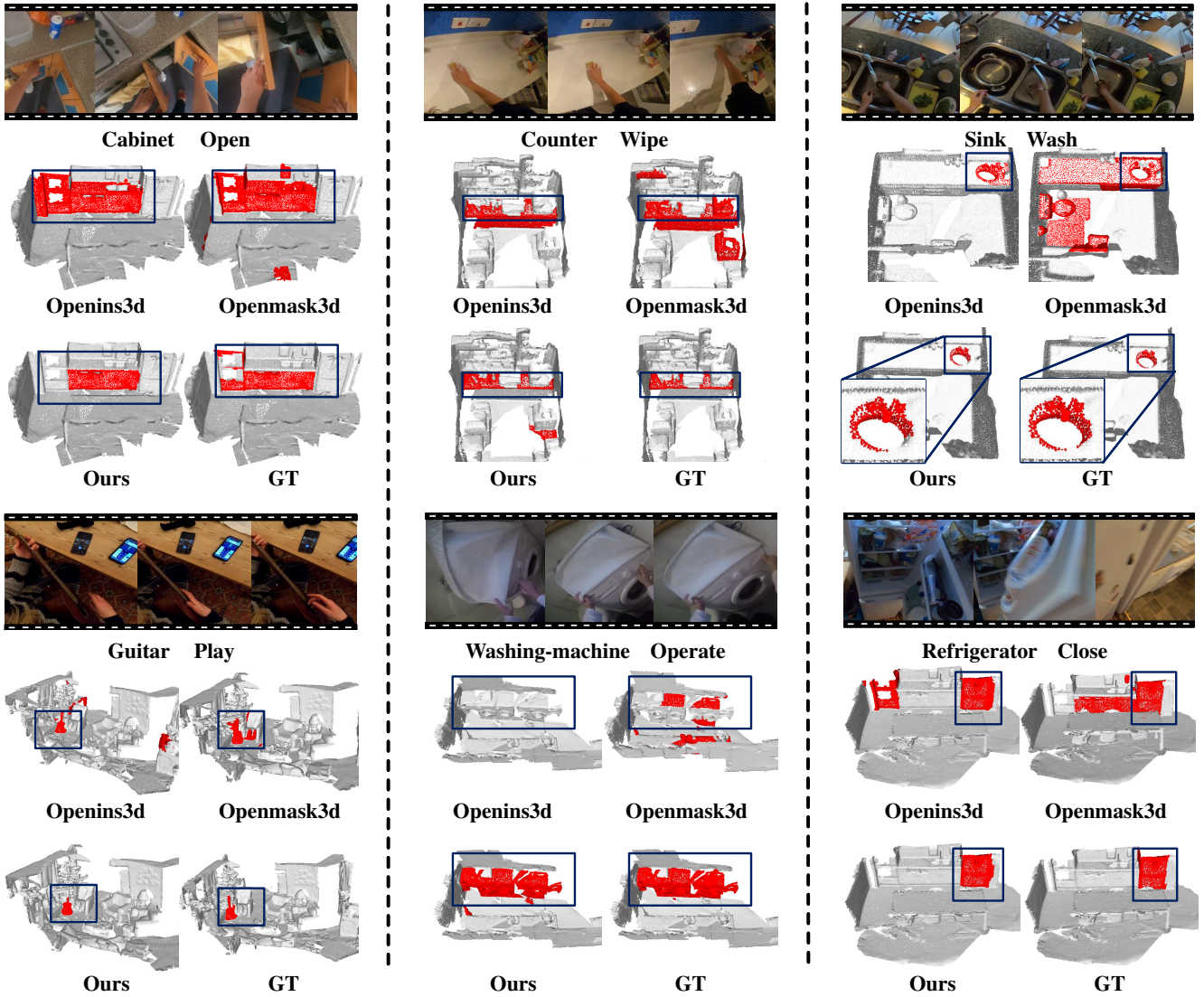

\textbf{Collection Details.\ } We constructed the \textbf{V}ideo-3D \textbf{S}cene \textbf{A}ffordance \textbf{D}ataset (VSAD), which comprises paired egocentric video and 3D scene affordance data. The egocentric interaction videos are primarily sourced from the Ego4D \cite{grauman2022ego4d} and EPIC-100 \cite{damen2022rescaling} datasets, while the 3D indoor scenes are derived from the leading 3D scene understanding datasets such as ScanNetV2\cite{dai2017scannet} and Matterport3D\cite{chang2017matterport3d}. The selection criteria are detailed as follows: the egocentric videos must explicitly depict interactions between humans and objects within a 3D scene. For instance, if the 3D scene includes an object like a ``chair'' with the affordance ``move,'' the video must demonstrate the process of a person moving the chair from an egocentric perspective. Ambiguous and outdoor videos are removed by hand, while videos with incorrect annotations are manually corrected. Ultimately, we compiled 3,814 interaction video clips covering 17 affordance categories and 16 interaction targets. All scenes from ScanNetV2 are included. To mitigate the long-tail effect and enhance dataset diversity, we select and add additional scenes from Matterport3D as supplements. The final dataset comprises 2,086 3D indoor scenes, encompassing 7,690 interactive areas corresponding to video interactions. The training set consists of 3,063 video clips and 1,649 3D scenes, while the validation set includes 751 video clips and 437 3D scenes. The accompanying Fig. \ref{Fig:Dataset} (a) shows paired data examples.

\textbf{Statistic Analysis.\ } To provide a more comprehensive and thorough understanding of the \textbf{VSAD} dataset, we present its statistical characteristics in Fig. \ref{Fig:Dataset}, which offers valuable and insightful information about the dataset’s distribution. It is important to emphasize that the egocentric videos and scenes are sampled from different sources, meaning there is no requirement for a strict one-to-one correspondence between them. In practical, a single 3D scene can be linked to multiple egocentric video clips, with the number of pairings varying based on the complexity and diversity of the interactions within each scene. Fig. \ref{Fig:Dataset} (b) presents a detailed analysis of the number and distribution of objects across various affordance categories within the video clips, underscoring the diversity of affordance and objects, and highlighting the dataset’s ability to capture a wide range of interactions. Furthermore, Fig. \ref{Fig:Dataset} (c) illustrates the proportion of videos and scenes associated with each object category, providing key insights into the balance between video content and 3D scene representation, which further demonstrates the dataset’s comprehensiveness.

\section{Experiment}
\subsection{Benchmark Setting}
\textbf{Evaluation Metrics.\ }To provide a comprehensive and effective evaluation, we integrate several advanced achievements in the fields of affordance grounding and 3D scene understanding and selecte the following evaluation metrics: mAP, AP50, AP25, mRC, RC50, and RC25. Specifically, mean Average Precision (mAP) and mean Recall (mRC) are calculated by averaging the scores over IoU thresholds ranging from 50\% to 95\% in 5\% increments. AP50 and RC50 represent the scores at an IoU threshold of 50\%, while AP25 and RC25 correspond to an IoU threshold of 25\%. These metrics effectively assess the precision and coverage of affordance prediction masks across different IoU thresholds.

\begin{table*}[t]
\small
\renewcommand{\arraystretch}{1.} 
\centering
  \caption{{\textbf{Quantitative Results.} The overall results of all comparative methods, the best results are in \textbf{bold} and covered with the grey mask. ``\textbf{Base}.'' represents the baseline. ``\textbf{Ours.I}'' represents that the 2D input is a single frame. $\textcolor{darkpink}{\scriptstyle~\diamond}$ indicates the relative improvement to the first row. All results are shown in percentage.}}
\label{table:results}
\begin{tabular}{c|cccccc}
\toprule
\textbf{Method}   & \textbf{mAP $(\uparrow)$} & \textbf{AP50 $(\uparrow)$} & \textbf{AP25 $(\uparrow)$} & \textbf{mRC $(\uparrow)$} & \textbf{RC50 $(\uparrow)$} & \textbf{RC25 $(\uparrow)$}\\ \midrule[1.2pt]
\textbf{Base.}  &$3.484$    & $6.664$       &   $10.537$      & $22.911$     & $30.367$  & $35.742$ \\
\textbf{Open3dis\cite{nguyen2024open3dis}}  &$4.425\textcolor{darkpink}{\scriptstyle~\diamond27\%}$    & $6.021\textcolor{darkpink}{\scriptstyle~\diamond-9\%}$       &   $8.101\textcolor{darkpink}{\scriptstyle~\diamond-23\%}$      & $24.701\textcolor{darkpink}{\scriptstyle~\diamond8\%}$     & $31.584\textcolor{darkpink}{\scriptstyle~\diamond4\%}$  & $37.533\textcolor{darkpink}{\scriptstyle~\diamond5\%}$ \\
\textbf{Openmask3d\cite{takmaz2023openmask3d}} &$5.059\textcolor{darkpink}{\scriptstyle~\diamond45\%}$    & $6.989\textcolor{darkpink}{\scriptstyle~\diamond5\%}$       &   $8.623\textcolor{darkpink}{\scriptstyle~\diamond-18\%}$      & $29.943\textcolor{darkpink}{\scriptstyle~\diamond31\%}$     & $35.378\textcolor{darkpink}{\scriptstyle~\diamond17\%}$  & $38.259\textcolor{darkpink}{\scriptstyle~\diamond7\%}$ \\
\textbf{Openins3d\cite{huang2023openins3d}}  &$8.348\textcolor{darkpink}{\scriptstyle~\diamond139\%}$    & $10.000\textcolor{darkpink}{\scriptstyle~\diamond50\%}$       &   $10.562\textcolor{darkpink}{\scriptstyle~\diamond0.2\%}$      & $25.001\textcolor{darkpink}{\scriptstyle~\diamond9\%}$     & $35.965\textcolor{darkpink}{\scriptstyle~\diamond18\%}$  & $36.222\textcolor{darkpink}{\scriptstyle~\diamond1\%}$ \\ \midrule[1.2pt]
\textbf{Ours.I}  &$6.307\textcolor{darkpink}{\scriptstyle~\diamond81\%}$    & $10.208\textcolor{darkpink}{\scriptstyle~\diamond54\%}$       &   $13.599\textcolor{darkpink}{\scriptstyle~\diamond29\%}$      & $32.664\textcolor{darkpink}{\scriptstyle~\diamond43\%}$     & $46.452\textcolor{darkpink}{\scriptstyle~\diamond53\%}$  & $57.204\textcolor{darkpink}{\scriptstyle~\diamond60\%}$ \\
\rowcolor{mygray}
\textbf{Ours}  & \textbf{10.652}$\textcolor{darkpink}{\scriptstyle~\diamond203\%}$    &\textbf{18.243}$\textcolor{darkpink}{\scriptstyle~\diamond173\%}$      & \textbf{24.036}$\textcolor{darkpink}{\scriptstyle~\diamond128\%}$      &  \textbf{36.829}$\textcolor{darkpink}{\scriptstyle~\diamond60\%}$     & \textbf{52.616 }$\textcolor{darkpink}{\scriptstyle~\diamond73\%}$  & \textbf{66.523 }$\textcolor{darkpink}{\scriptstyle~\diamond86\%}$               \\ 
\bottomrule[1.2pt]
\end{tabular}
\end{table*}

\textbf{Modular Baselines.\ }Similar research to our methodological paradigm typically uses text to locate target regions in a scene. To highlight the advantages of our approach, we compare it with several state-of-the-art open-vocabulary 3D scene understanding methods \cite{takmaz2023openmask3d,nguyen2024open3dis,huang2023openins3d}.These methods perform well on instance segmentation task and can map semantics to specific structures using an additional text encoder, enabling region grounding in the scene through open-vocabulary queries such as object category, color, material, and other characteristics. For consistency, we use the category names of the target objects interacted within the videos as queries. For the baseline, we remove the \textbf{ISA} and \textbf{BQD} modules, treat a cross-attention layer before the decoder as a cross-fusion module, and add an MLP to predict quality scores. Furthermore, to demonstrate that egocentric videos with richer interaction clues are more effective for learning scene affordance than a single image, we use only a center frame from the video as the 2D input and employ ResNet50 to extract image features.

\begin{figure}[t]
	\centering
	\small
        \begin{overpic}[width=1.\linewidth]{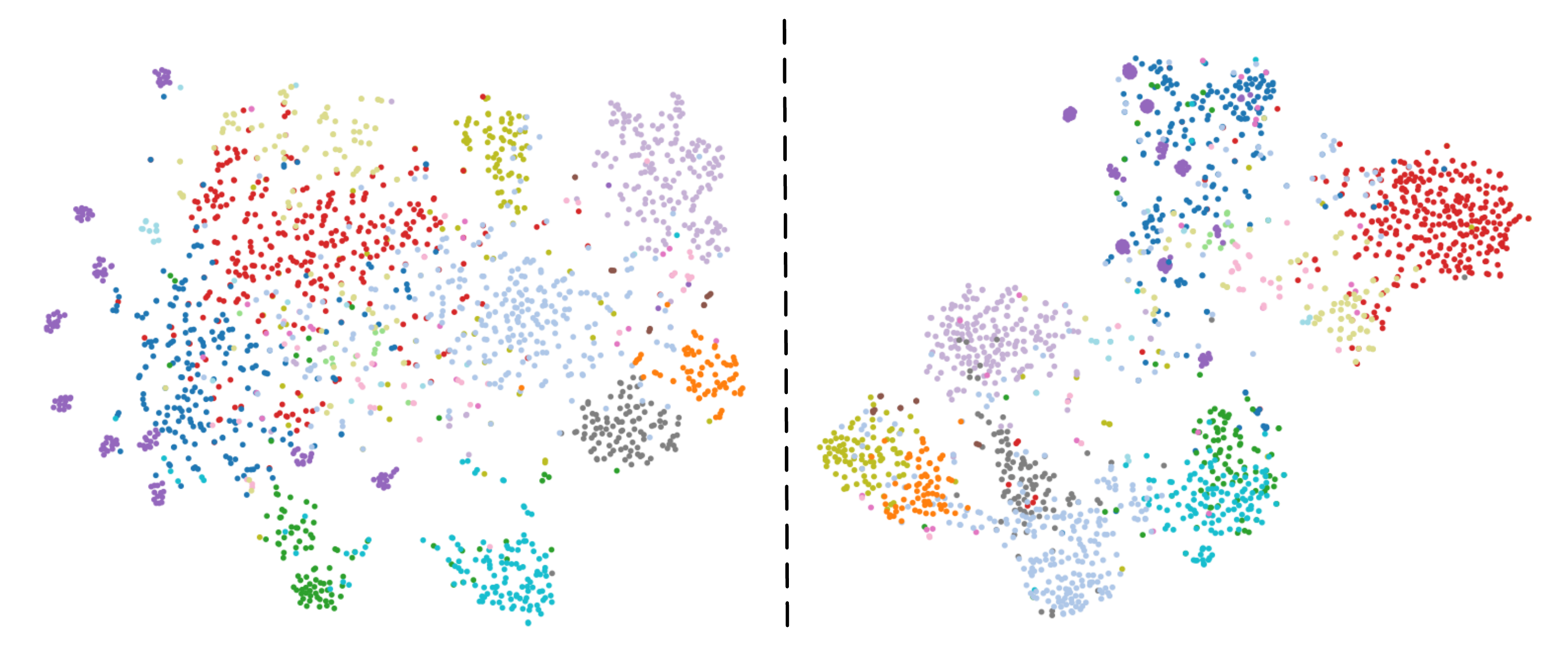}
        \put(25,0){{\textbf{(a)}}}
        \put(72,0){{\textbf{(b)}}}
	\end{overpic}
 \vspace{-10pt}
	\caption{\textbf{T-SNE Visualization Results.} The t-SNE results for the baseline without any modules \textbf{(a)} and our model \textbf{(b)}. }
 \label{Fig:t-sne}
\end{figure}

\begin{table}
\centering
\small
\renewcommand{\arraystretch}{1.}
\renewcommand{\tabcolsep}{5.0pt}
  \caption{{\textbf{Ablation Study on Framework Design.} Metrics when detaching the $\textbf{ISA}$, $\textbf{BQD}$ module, and the progressive prediction head ($\textbf{P.P.}$). $\textbf{\emph{w/o}}$ means without.}}
\label{table:ablation1}
\vspace{1pt}
\begin{tabular}{c|cccccc}
\toprule
\textbf{Metrics}   & \textbf{mAP }  & \textbf{AP50 } & \textbf{AP25 } & \textbf{mRC } & \textbf{RC50 } & \textbf{RC25 } \\ \midrule
$\textbf{\emph{w/o} ISA}$  &$8.241$    & $14.722$       &   $20.651$      & $33.324$     & $50.183$  & $63.671$ \\
\textbf{\emph{w/o} BQD}  &$7.313$    & $13.019$       &   $19.232$      & $29.182$     & $46.004$  & $62.059$ \\
\textbf{\emph{w/o} P.P.}  &$6.342$    & $12.025$       &   $19.233$      & $29.885$     & $46.380$ & $64.803$\\
\rowcolor{mygray}
\textbf{Ours}  &$\textbf{10.652}$    & $\textbf{18.243}$       &   $\textbf{24.036}$      & $\textbf{36.829}$     & $\textbf{52.616}$  & $\textbf{66.523}$ \\ \bottomrule
\end{tabular}
\end{table}
\textbf{Implementation Detial.\ }Our model is implemented in PyTorch and trained with the AdamW optimizer\cite{kingma2014adam}. The input videos are downsampled to 16 frames and each frame is randomly clipped from 256 × 256 to 224 × 224 with random horizontal flipping. For the input 3D scenes, we first extract its superpoints using the method in \cite{landrieu2018large}, and then preprocess it using the same method as ScanNetv2. We train the model for 250 epochs on four NVIDIA RTX3090 GPUs with an initial learning rate of $1e^{-4}$. The hyperparameters $\lambda_1$, $\lambda_2$, $\lambda_3$, and $\lambda_4$ are set to 1, 0.5, 0.5, and 0.5, respectively, with $L$ and $N$ set to 6 and 50.

\subsection{Comparison Results}
\textbf{Quantitative Results.\ }The results of the objective metrics, as presented in Tab. \ref{table:results}, demonstrate that our method consistently outperforms all other approaches across all evaluated metrics. The baseline model, which produces poor results, suggests that simple cross-attention mechanism is insufficient to capture the intricate correlations between 2D and 3D sources. Notably, our approach exceeds all methods relying on fixed semantic-geometric mappings to anchor corresponding regions. For instance, compared to Openins3d, our method achieves improvements of $\bm{+2.304}$ in \textbf{mAP}, $\bm{+8.243}$ in \textbf{AP50} and $\bm{+13.474}$ in \textbf{AP25}. By exploring the relationship between interaction intent and scene layout, our method effectively reduces false affordance mask predictions in non-interactive areas. Furthermore, the use of the \textbf{BQD} mechanism further enhances the alignment of affordance clues between different modalities, contributing to the superior performance of our model. The results from the model trained on single interaction images also highlight the difficulty of capturing complex interactions within a single frame, further validating the superiority of our design.

\textbf{Qualitative Results.\ }We present a visual comparison of the affordance maps generated by our method and other approaches, as illustrated in Fig. \ref{Fig:result}. Our method consistently excels at precisely identifying affordance regions within a scene. For instance, our method accurately grounds the most relevant affordance region for the interaction, while the \textbf{ISA} module effectively ignores irrelevant areas. In more complex scenes, such as the fifth example ``operate washing machine,'' where multiple regions offer the same affordance, our approach produces a more comprehensive affordance maps. Unlike methods that rely on fixed semantic-structure mappings, our approach adeptly predicts scene affordance by considering not only geometry but also interaction properties, avoiding confusion with objects of similar structure. These results reinforce the validity of our method and highlight its superiority.

\begin{table*}[t]
\begin{minipage}[m]{0.24\textwidth}
    \caption{\textbf{Ablation Study on Loss Fuction Design.} We investigate the impact of the $\mathcal{L}_{con}$, $\mathcal{L}_{kl}$ and $\mathcal{L}_{CE}$.}
     \label{table:ablation2}
\end{minipage}
\begin{minipage}[t]{0.75\textwidth}
\centering
\small
  \renewcommand{\arraystretch}{1.}
  \renewcommand{\tabcolsep}{6.5pt}
    \begin{tabular}{ccc|cccccc}
    \toprule[1.2pt]
      $\textbf{$\mathcal{L}_{con}$}$  &    $\textbf{$\mathcal{L}_{kl}$}$   &   $\textbf{$\mathcal{L}_{CE}$}$&   \textbf{mAC} $(\uparrow)$ & \textbf{AC50} $(\uparrow)$ & \textbf{AC25} $(\uparrow)$ &\textbf{ mRC} ($\uparrow$) & \textbf{RC50} ($\uparrow$)  & \textbf{RC25} ($\uparrow$)     \\
        \midrule[1.2pt]
                   &              &    \cmark &    $5.819$  & $10.206$   & $14.062$ & $33.095$ & $45.734$ & $57.993$ \\
                   &  \cmark      &   \cmark  &    $7.932$ & $14.348$ & $19.166$ & $33.533$ & $47.742$ & $63.369$  \\
      \cmark       &              &   \cmark  &   $8.761$ & $16.117$ & $21.394$ & $35.595$ & $51.477$ &  $65.308$     \\
         \cmark    &  \cmark        &          &    $9.292$ & $16.588$ & $18.983$ & $38.192$ & $54.623$ &  $65.950$     \\
      \rowcolor{mygray}
      \cmark       &  \cmark       &   \cmark &  $\textbf{10.652}$ & $\textbf{18.243}$ & $\textbf{24.036}$ & $\textbf{36.829}$ & $\textbf{52.616}$ &  $\textbf{66.523}$  \\
        \bottomrule[1.2pt]
    \end{tabular}
\end{minipage}
\end{table*}

\begin{figure}
	\centering
	\small
        \begin{overpic}[width=1.\linewidth]{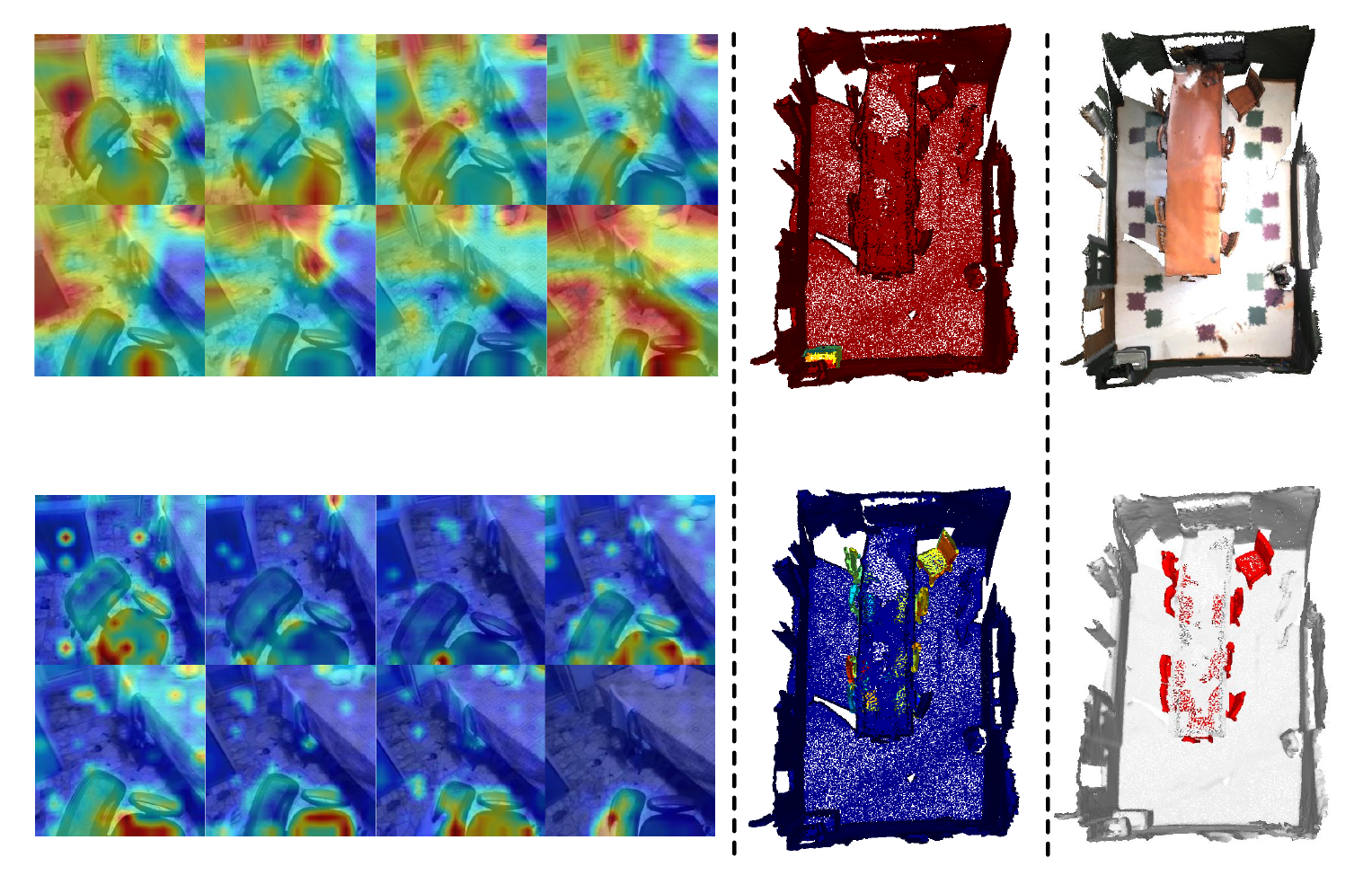}
        \put(17.5,32.5){{\textbf{Chair}}}
        \put(28.5,32.5){{\textbf{Move}}}
        \put(26,-1){{\textbf{(a)}}}
        \put(63,-1){{\textbf{(b)}}}
        \put(86,32){{\textbf{(c)}}}
        \put(86,-1){{\textbf{(d)}}}
	\end{overpic}
 \vspace{-10pt}
    \caption{\textbf{Saliency Map.} \textbf{(a)} Visualization of the relevance map from $Q_v$ to original video and \textbf{(b)} $Q_s$ to original 3D scene in \textbf{BQD} module. The first row is for queries in layer 1, and the second row is for queries in layer 6. \textbf{(c)} The original input 3D scene. \textbf{(d)} The affordance grounding results. }
 \label{Fig:attention_map}
\end{figure}
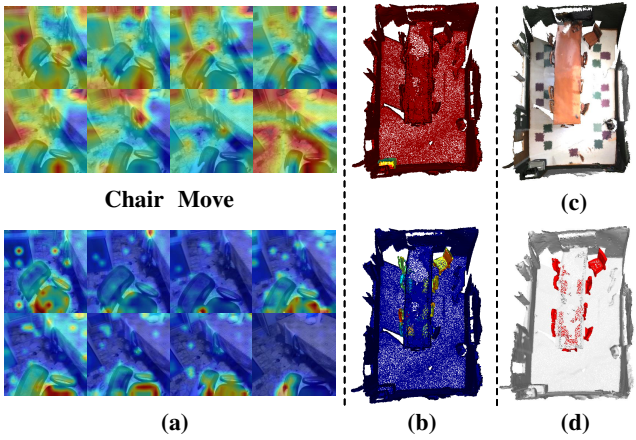 

\subsection{Ablation Study}
In this section, we conduct a comprehensive ablation study to investigate the effect of different framework design, loss function design and hyper-parameter settings.

\textbf{Framework Design.\ }The framework ablation results are shown in Tab. \ref{table:ablation1}. It reports the model performance without \textbf{ISA}, which captures the correlations of interaction intent and sub-region layout. Without this module, the network struggles to identify significant interaction regions, resulting in incorrect predictions in irrelevant areas and a marked decline in both performance and efficiency. Similarly, excluding \textbf{BQD} hinders the model’s ability to extract and align affordance contexts across modalities, leading to a significant drop in overall performance. We also evaluate the impact of the prediction head, which generates predictions at each layer of \textbf{BQD}. Removing these heads and computing the loss solely from the final prediction disrupts progressive supervision, making it harder for the model to converge and significantly reducing its overall performance. To visually assess the effectiveness of our design, we employ t-SNE \cite{van2008visualizing} to illustrate the clustering of affordance for the baseline model (without any modules) and our enhanced model, as shown in Fig. \ref{Fig:t-sne}. The results indicate that our method establishes more distinct discriminative boundaries, enabling clearer differentiation of various affordance features. Additionally, we visualize the relevance maps of $\mathbf{Q}_v$ to the original video and $\mathbf{Q}_s$ to the original scene at layer 1 and layer 6 in the $\textbf{BQD}$ module. The visualizations reveal that as the layers deepen, $\mathbf{Q}_v$ and $\mathbf{Q}_s$ increasingly focus on affordance-related contexts in the video and scene features, confirming the validity and effectiveness of our module design.

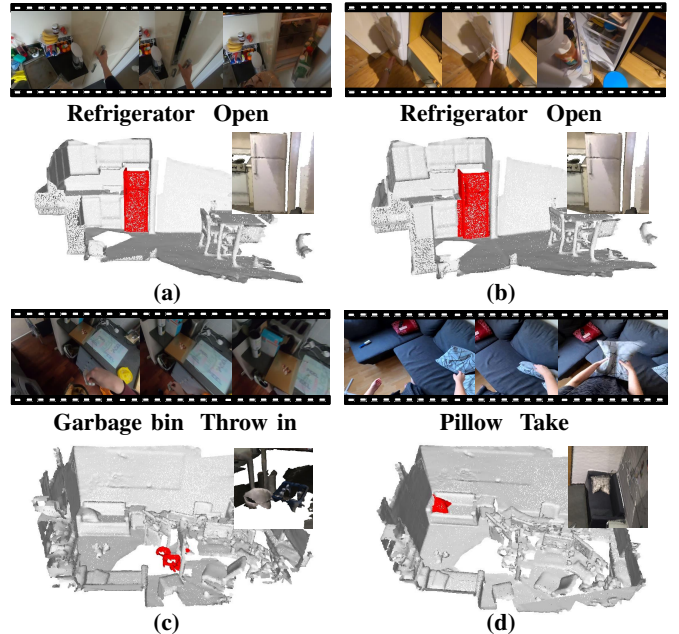
\begin{figure}
	\centering
	\small
        \begin{overpic}[width=1.\linewidth]{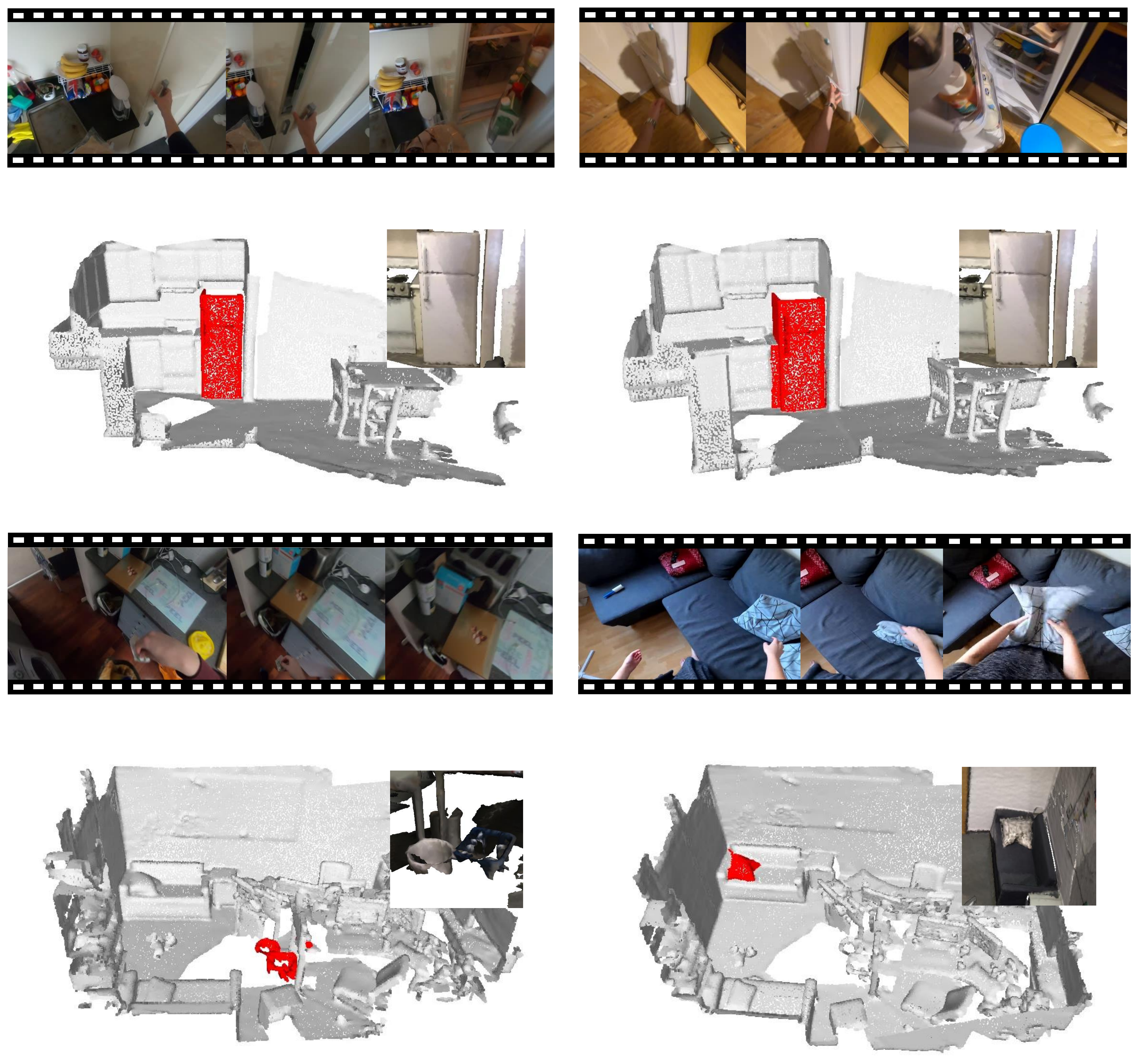}
        \put(9,75){{\textbf{Refrigerator}}}
        \put(31,75){{\textbf{{Open}}}}
        \put(59,75){{\textbf{Refrigerator}}}
        \put(81,75){{\textbf{Open}}}
        
        \put(7,28.5){{\textbf{Garbage bin}}}
        \put(29,28.5){{\textbf{Throw in}}}

        \put(65,28.5){{\textbf{Pillow}}}
        \put(77,28.5){{\textbf{Take}}}
        
        \put(22,48){{\textbf{(a)}}}
        \put(72,48){{\textbf{(b)}}}
        \put(22,-1.5){{\textbf{(c)}}}
        \put(72,-1.5){{\textbf{(d)}}}
	\end{overpic} 
    \caption{\textbf{Performance Analysis.} Grounding affordance on the same 3D scene with videos that demonstrate the same/different interactions. \textbf{(a)}, \textbf{(b)}: Different videos with the same affordance $w.r.t.$ Single scene; \textbf{(c)}, \textbf{(d)}: Different videos with different affordance $w.r.t.$ Single scene. }
 \label{Fig:analysis1}
\end{figure}

\textbf{Loss Function Design.\ }We investigate the impact of different loss functions, with the results presented in Tab. \ref{table:ablation2}. Since other loss functions have minimal effect without the supervision of $\mathcal{L}_{mask}$, all results are from experiments that include $\mathcal{L}_{mask}$. It indicates that using $\mathcal{L}_{CE}$ implicitly optimizes the alignment process, enhancing model performance. Notably, the improvement achieved by $\mathcal{L}_{con}$ is more significant than that of $\mathcal{L}_{KL}$, suggesting that enhancing the separation between positive and negative examples more effectively boosts the model’s ability to perceive and identify distinct affordance regions. Additionally, the incorporation of $\mathcal{L}_{KL}$, compared to using only $\mathcal{L}_{CE}$, substantially improves prediction accuracy by constraining the model to align affordance-related features more precisely. The above results highlight the effectiveness of our loss function design.

\begin{figure*}
	\centering
	\small
        \begin{overpic}[width=1.\linewidth]{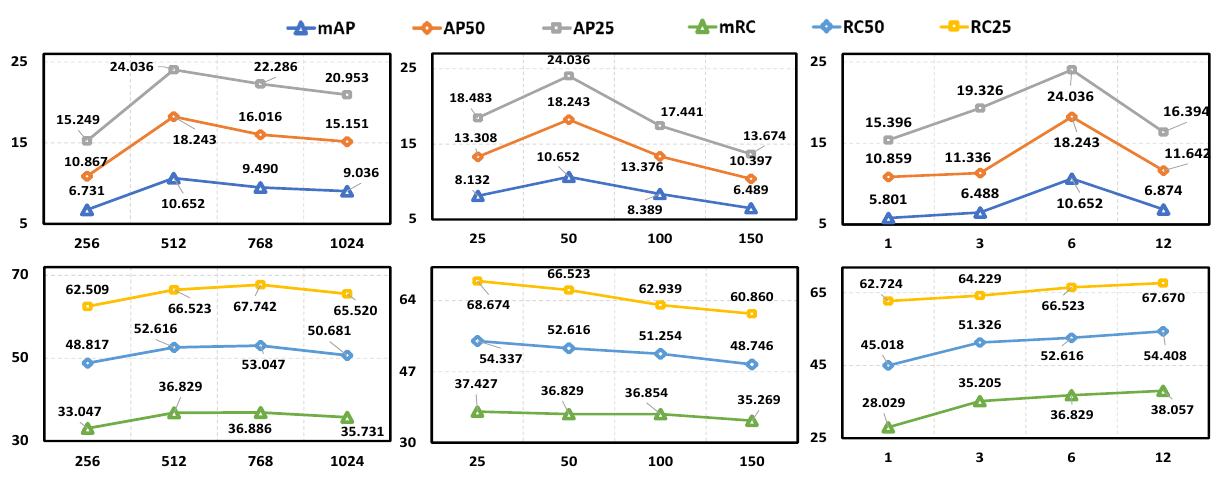}
        \put(17,-1){{\textbf{(a)}}}
        \put(49,-1){{\textbf{(b)}}}
        \put(82.5,-1){{\textbf{(c)}}}
	\end{overpic}
 \vspace{-10pt}
    \caption{\textbf{Different Hyper-parameters.} We explore \textbf{(a)} the effects of different model dimension $C$, \textbf{(b)} the effects of different number of layers in \textbf{BQD} $L$, and \textbf{(c)} the effects of different number of bilateral query vectors $N$.}
 \label{Fig:hyper_param}
\end{figure*}

\textbf{Hyper-parameter Settings.\ }We conduct experiments on several key hyper-parameters, including the model dimension $C$, the number of bilateral query vectors $N$, and the number of layers in \textbf{BQD} $L$. The results are presented in Fig. \ref{Fig:hyper_param}. As shown in Fig. \ref{Fig:hyper_param} (a), $C$ significantly impacts model precision and has a smoothing effect on coverage. A smaller $C$ may yield poor results, as limited dimensionality reduces the network's ability to express complex patterns. Conversely, excessively increasing $C$ may complicate optimization and degrade performance. A larger number of channels may increase the complexity of the optimization and lead to a decrease in model performance. Fig. \ref{Fig:hyper_param} (b) illustrates the effect of $N$, where each query corresponds to an affordance mask for the final prediction. The optimal performance is achieved at $N = 50$. Larger values of $N$ tend to worsen performance, likely due to query redundancy, with multiple queries focusing on the same affordance region or interaction-independent region, leading to an increase in false positive.
Fig. \ref{Fig:hyper_param} (c) demonstrates the impact of $L$, showing a generally positive effect on model performance as $L$ increases from 1 to 6. This trend suggests that the BQD module progressively captures affordance clues in 2D interactive videos and 3D scenes, which plays a critical role in affordance prediction. However, when $L$ becomes too large, the network becomes overly complex, reducing efficiency and increasing the rate of incorrect predictions. Finally, balancing efficiency and performance, we set $C$, $N$, and $L$ to 512, 50, and 6, respectively.

\subsection{Performance Analysis}
\textbf{Different Videos with Same Affordance} $w.r.t.$ \textbf{Single Scene}: In real-world scenarios, multiple videos may depict the same affordance despite variations in interaction styles, object appearance and background environment. Consequently, our model is designed to consistently identify the same affordance regions within identical 3D scenes, regardless of these variations. As illustrated in Fig. \ref{Fig:analysis1} (a) and (b), despite being described by different videos, the affordance region in the corresponding 3D scene is correctly detected. These results demonstrate that our model is robust to these variations and effectively captures the common affordance knowledge shared among the videos.

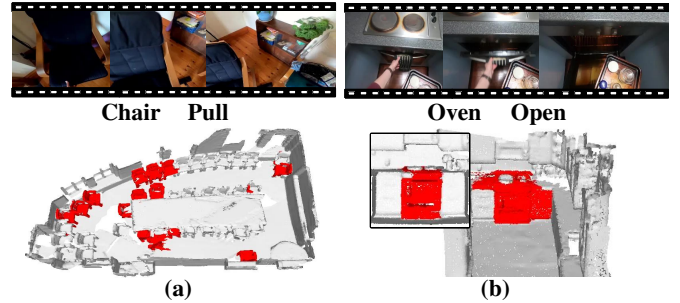
\begin{figure}
	\centering
	\small
        \begin{overpic}[width=1.\linewidth]{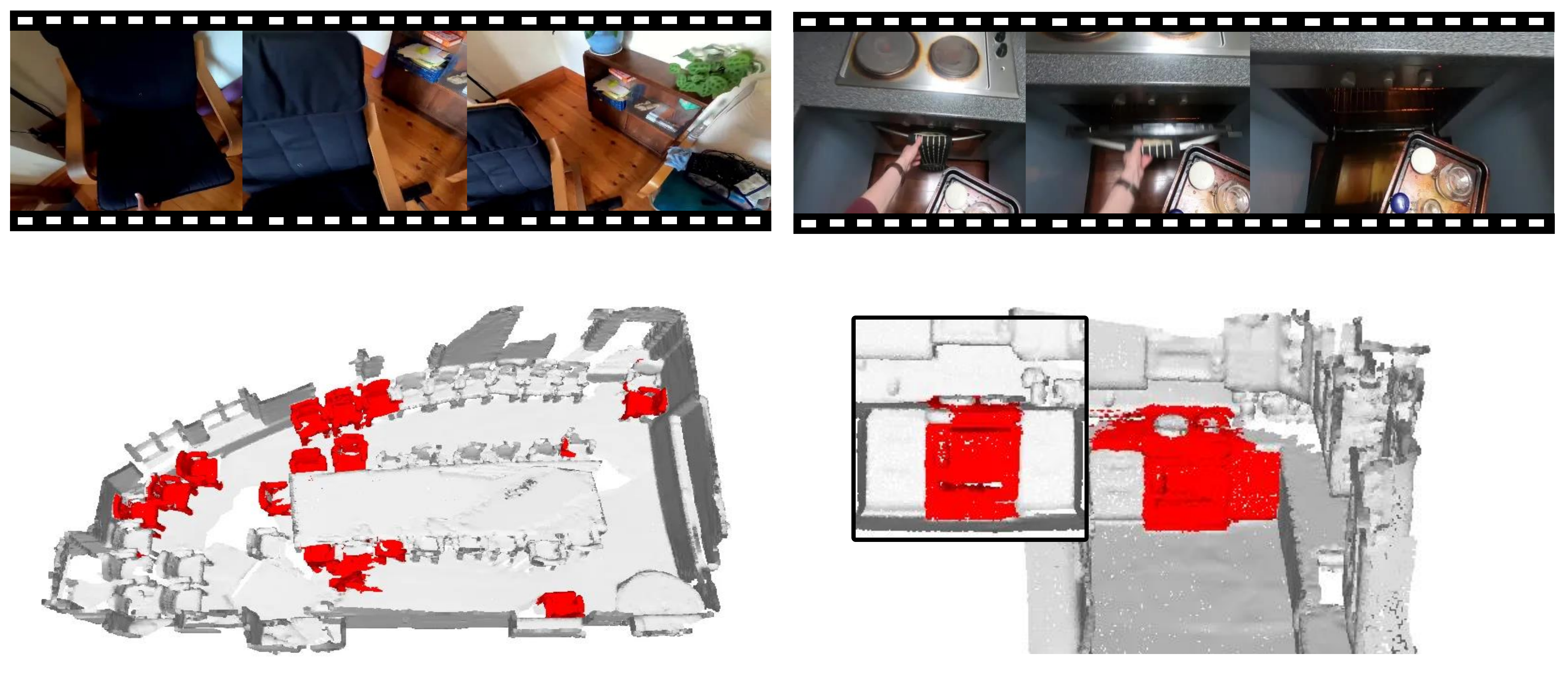}
        \put(14,25){{\textbf{Chair}}}
        \put(27,25){{\textbf{Pull}}}
        \put(63,25){{\textbf{Oven}}}
        \put(75.5,25){{\textbf{Open}}}

        \put(23.5,-1.5){{\textbf{(a)}}}
        \put(71,-1.5){{\textbf{(b)}}}
	\end{overpic}
 \vspace{-10pt}
    \caption{\textbf{Failure Cases.} \textbf{(a)} Incomplete prediction in complex scene. \textbf{(b)} Over prediction in the area with blurred boundaries.}
 \label{Fig:failure case}
\end{figure}

\textbf{Different Videos with Different Affordance} $w.r.t.$ \textbf{Single Scene.} To better understand the surrounding environment, when given multiple interaction videos with different affordance, our model is expected to correctly localize the corresponding affordance regions in the 3D scene according to the specific interaction depicted in each video. Examples in Fig. \ref{Fig:analysis1} (c) and (d) show that our model accurately aligns the 2D interaction clues with 3D scene information, even when the affordance differs across videos. This further illustrates its capability to adapt to varying interaction clues while maintaining precise scene interpretation.

\textbf{Limitations.\ }Our model has certain limitations, shown in Fig . \ref{Fig:failure case}.  In complex 3D environments with numerous regions offering similar affordance,  the fixed query number strategy employed in the \textbf{BQD} module may restrict the maximum number of affordance masks that can be predicted. Furthermore, when affordance regions overlap with other objects, the method may struggle to delineate boundaries accurately, leading to over-prediction. This issue likely stems from the model’s limited capacity to capture fine-grained geometric details. To address these challenges, we plan to explore dynamic and adaptive strategies, as well as more advanced geometric modeling techniques in future work.
\section{CONCLUSION AND DISCUSSION} 
In this paper, we address the complex task of grounding 3D scene affordance from egocentric interactions, providing valuable interactive insights into the environment that can enhance embodied intelligence and applications like AR and VR. We first introduce a new dataset \textbf{VSAD}, curated from existing datasets, which pairs egocentric video with 3D scene affordance data, creating a pioneering benchmark for this task. Building on this, we propose a novel framework \textbf{Ego-SAG}, designed to extract and align interaction information from multiple sources to predict scene affordance accurately. Specifically, the \textbf{ISA} module captures the relationship between interaction intent and sub-region layout, focusing on areas most relevant to a specific interaction, the \textbf{BQD} module constructs a dynamic affordance map across modalities through a bilateral query decoder mechanism, progressively extracting affordance features and optimizing high-dimensional alignment to precisely identify 3D scene affordance. Compared to existing approaches, our model surpasses performance across all evaluation metrics, offering novel insights and advancing the field of affordance learning.

\bibliographystyle{IEEEtran}

\clearpage
\end{document}